%% file: main.tex
\documentclass[conference]{IEEEtran}
\IEEEoverridecommandlockouts

\usepackage{cite}
\usepackage{amsmath,amssymb,amsfonts}
\usepackage{graphicx}
\usepackage{textcomp}
\usepackage[table]{xcolor}

\usepackage[utf8]{inputenc}
\usepackage[english]{babel}

\usepackage{mathtools,amsthm,mathrsfs}
\usepackage{units}          
\usepackage{stmaryrd}       
\usepackage{dsfont}
\usepackage{bm}

\usepackage{algorithm}
\usepackage{algpseudocode}

\usepackage{tikz}
\usetikzlibrary{arrows.meta, calc, positioning, fit, decorations.pathreplacing, shapes.geometric}
\usepackage{pgfplots}
\pgfplotsset{compat=1.18}
\usepgfplotslibrary{fillbetween}
\usepackage[caption=false,font=footnotesize]{subfig}
\usepackage{stfloats}
\usepackage{subfig}
\usepackage{multirow, booktabs, tabularx, array}
\newcolumntype{C}[1]{>{\centering\arraybackslash}m{#1}}

\usepackage[hidelinks]{hyperref}
\usepackage[final]{microtype}

\theoremstyle{definition}

\theoremstyle{plain}

\theoremstyle{remark}

\newcommand*{\VEC}[1]{\ensuremath{\boldsymbol{#1}}}
\newcommand*{\MAT}[1]{\ensuremath{\boldsymbol{#1}}}

\DeclareMathOperator*{\argmin}{argmin}

\newcommand{\R}{\mathbb{R}}

\newcommand{\grad}{\mathrm{grad}}

\newcommand{\St}{\textrm{St}}
\DeclareMathOperator{\tr}{tr}
\DeclareMathOperator{\diff}{D}

\def\BibTeX{{\rm B\kern-.05em{\sc i\kern-.025em b}\kern-.08em
    T\kern-.1667em\lower.7ex\hbox{E}\kern-.125emX}}

\input{eegnet_figures.tex}
\begin{document}
\bstctlcite{IEEEexample:BSTcontrol}
\title{FedSPDnet: Geometry-Aware Federated Deep Learning with SPDnet}

\author{
\IEEEauthorblockN{Thibault Pautrel\textsuperscript{1}, Florent Bouchard\textsuperscript{1}, Ammar Mian\textsuperscript{2}, Guillaume Ginolhac\textsuperscript{2}}
\IEEEauthorblockA{\textsuperscript{1} CentraleSup\'{e}lec, L2S, France}
\IEEEauthorblockA{\textsuperscript{2}Universit\'{e} Savoie Mont-Blanc, LISTIC, France}

}

\maketitle

\begin{abstract}
We introduce two federated learning frameworks for the classical SPDnet model operating on symmetric positive definite (SPD) matrices with Stiefel-constrained parameters. Unlike standard Euclidean averaging, which violates orthogonality, our approach preserves geometric structure through two efficient aggregation strategies: ProjAvg, projecting arithmetic means onto the Stiefel manifold, and RLAvg, approximating tangent-space averaging via retractions and liftings. Both methods are computationally efficient, independent of the optimizer, and enable federated learning for signal processing applications whose features are SPD matrices. Simulations on EEG motor imagery benchmarks show that FedSPDnet outperforms federated EEGnet in F1 score and robustness to federation and partial participation, while using fewer parameters per communication round.
\end{abstract}


\section{Introduction}

Federated learning (FL) enables collaborative model training without centralizing raw data, typically by iteratively averaging parameters on a central server as in FedAvg~\cite{mcmahan2017communication,yu2019parallel,li2020federated}. Subsequent refinements such as FedProx~\cite{li2020federated,yuan2022convergence} and SCAFFOLD~\cite{karimireddy2020scaffold} improve robustness under data heterogeneity and client drift, but remain limited to Euclidean parameter spaces and do not extend to non-Euclidean geometries.

In parallel, the last decade has seen a rising interest in Riemannian optimization, which generalizes classical optimization to parameters living on manifolds~\cite{absil2008optimization,boumal2023introduction}. Such techniques are needed for learning problems with geometric constraints, including low-rank matrix factorization, dimensionality reduction, and orthogonal parameterizations. In deep learning, Riemannian models have proven particularly effective when operating on structured data such as symmetric positive definite (SPD) covariance matrices, leading to architectures like SPDnet~\cite{huang2017riemannian}. SPDnet integrates bilinear mappings on the Stiefel manifold with nonlinear eigenvalue rectification, ensuring that intermediate representations remain SPD while reducing dimensionality in a geometry-preserving way. This architecture has proven its effectiveness in several applications such as micro-Doppler radar~\cite{brooks2019riemannian}, electroencephalography~\cite{kobler2022spd}, and ground-penetrating radar~\cite{jafuno2025classification}.

Some works~\cite{li2022federated,zhang2024nonconvex,huang2026federated,huang2024riemannian} have proposed adapting federated learning architectures to general Riemannian manifolds and do not concern the SPDnet network. In this paper, we therefore propose new federated architectures that assume that all clients use the SPDnet network. To develop a manifold-based federated learning model, the main challenge lies in aggregating parameters: while local training on clients leverages optimization methods adapted to the manifold, the server-side averaging step must also be designed to obtain a result that belongs to the manifold. The optimal solution is therefore to use a Riemannian mean~\cite{boumal2023introduction}, which allows to obtain the barycenter of a set of parameters belonging to the specific manifold. Unfortunately, this step is problematic for SPDnet: on the Stiefel manifold, where its orthogonal weights live, the Riemannian logarithm and geodesic distance have no closed form, so the Riemannian mean must be computed by iterative optimization~\cite{bouchard2025beyond}, making it too costly to run at every communication round and for every layer.

Inspired by the geometric tools from~\cite{bouchard2025beyond}, we propose two closed-form, geometry-preserving aggregation strategies for SPDnet: (i)~ProjAvg, which computes Euclidean averages of local weights and maps them back onto the Stiefel manifold via polar decomposition, and (ii)~RLAvg, which approximates tangent-space Riemannian averaging using retractions and liftings. Both methods are simple to implement and preserve the Stiefel constraint with only a single projection (or retraction--lifting step) per layer, enabling federated deployment of SPDnet without the cost of computing an exact Riemannian mean. In particular, the proposed architecture is more general than the one proposed in~\cite{huang2024riemannian} because it does not depend on the local optimizer used in the clients. Numerical experiments on real EEG data show that our federated models outperform federated EEGnet and retain more of the centralized performance than this Euclidean baseline.

\section{Background}

We address a supervised classification problem in which each data point is an EEG trial represented by its spatial covariance matrix $\MAT{\Sigma}\in\mathcal{S}^{++}_{d_0}$, the SPD matrix itself serving as the feature, with an associated label $y\in\llbracket1,K\rrbracket$. SPDnet learns a map from $\mathcal{S}^{++}_{d_0}$ to class scores.

This section presents the key elements that are needed to develop our proposed federated learning method adapted to the SPDnet architecture.
Since SPDnet weights are orthogonal, we first introduce Riemannian geometry and optimization on the Stiefel manifold.
Then, we provide details on the SPDnet architecture.

\subsection{Riemannian geometry and optimization on Stiefel}
\label{ssec:riem-backgd}
The Stiefel manifold~\cite{absil2008optimization,boumal2023introduction} is defined as $\St_{p,k}=\{\MAT{W}\in\mathbb{R}^{p\times k}: \, \MAT{W}^\top\MAT{W}=\MAT{I}_k\}$.
At $\MAT{W}\in\St_{p,k}$, the tangent space is $T_{\MAT{W}}\St_{p,k}=\{\MAT{X}\in\mathbb{R}^{p\times k}: \, \MAT{W}^\top\MAT{X}+\MAT{X}^\top\MAT{W}=\MAT{0}\}$.
We endow it with the ambient-space Frobenius metric: for $\MAT{X},\MAT{Y}\in T_{\MAT{W}}\St_{p,k}$, $\langle\MAT{X},\MAT{Y}\rangle_{\MAT{W}}=\tr(\MAT{X}^\top\MAT{Y})$.
From there, in the context of federated learning, we need geometrical tools on $\St_{p,k}$ to: 
\textit{(i)}~perform Riemannian optimization, and
\textit{(ii)}~aggregate points on $\St_{p,k}$.

Concerning the optimization of a cost function, one needs the Riemannian gradient to obtain a descent direction on the tangent space and a retraction to get a new iterate on the manifold from it.
Given a cost function $f:\St_{p,k}\to\mathbb{R}$, the Riemannian gradient $\grad f(\MAT{W})$ at $\MAT{W}$ is the unique tangent vector such that, $\forall\MAT{X}\in T_{\MAT{W}}\St_{p,k}$,
$
    \langle \grad f(\MAT{W}), \MAT{X} \rangle_{\MAT{W}}
    = \diff f(\MAT{W})[\MAT{X}],
$
where $\diff f(\MAT{W})[\MAT{X}]$ denotes the directional derivative of $f$ at $\MAT{W}$ in direction $\MAT{X}$.
For $\St_{p,k}$, it can be obtained from the Euclidean gradient $\nabla f(\MAT{W})$ through $\grad f(\MAT{W})=P_{\MAT{W}}(\nabla f(\MAT{W}))$, where $P_{\MAT{W}}:\mathbb{R}^{p\times k}\to T_{\MAT{W}}\St_{p,k}$ is the orthogonal projection~\cite{absil2008optimization,boumal2023introduction}
\begin{equation}
    P_{\MAT{W}}(\MAT{X}) = \MAT{X} - \frac{1}{2}\MAT{W}(\MAT{W}^\top\MAT{X}+\MAT{X}^\top\MAT{W}).
\label{eq:st_tangent_proj}
\end{equation}
On Stiefel, a convenient retraction $R_{\MAT{W}}:T_{\MAT{W}}\St_{p,k}\to\St_{p,k}$ at $\MAT{W}$ is the polar retraction~\cite{absil2008optimization,boumal2023introduction}
\begin{equation}
    R_{\MAT{W}}(\MAT{X}) = \mathcal{P}(\MAT{W}+\MAT{X}) = \mathrm{uf}(\MAT{W}+\MAT{X}),
\label{eq:st_retr}
\end{equation}
where $\mathcal{P}:\mathbb{R}^{p\times k}\to\St_{p,k}$ is the projection on Stiefel obtained by taking the orthogonal factor of the polar decomposition. This retraction returns the closest point on $\St_{p,k}$ to $\MAT{W}+\MAT{X}$ in Frobenius norm and admits a closed form.

To aggregate points on a Riemannian manifold, the natural choice is the Riemannian (Fr\'echet) mean, defined as the minimizer of the sum of squared geodesic distances to the input points, where geodesics generalize straight lines on the manifold. It rarely admits a closed form and is then computed iteratively by the so-called Karcher flow~\cite{karcher2014riemannian}, whose steps rely on the Riemannian exponential and logarithm~\cite{absil2008optimization,boumal2023introduction}. On $\St_{p,k}$, the logarithm and geodesic distance are not available in closed form and must be evaluated iteratively, so even a single step is costly, and a cheaper aggregation must be found.
It will be the contributions of Section~\ref{subsec:orthoagg}.

\subsection{SPDnet architecture}

\begin{figure*}[!t]
\centering
\resizebox{0.8\textwidth}{!}{%
\begin{tikzpicture}[
    >=stealth,
    font=\small,
    node distance=2.1cm,
    layer/.style={draw, thick, rounded corners, align=center,
                  minimum height=0.9cm, minimum width=1.2cm, fill=white},
    arrow/.style={->, thick, line cap=round, line join=round}
]

\node[layer, fill=white] (input) {Input\\$\MAT{\Sigma}_0 \in \mathcal{S}_{d_0}^{++}$} ;

\node[layer, right of=input] (bimap1) {\textbf{BiMap}\\$\MAT{W}_1$};
\draw[arrow] (input) -- (bimap1);
\node[layer, right of=bimap1] (re1) {\textbf{ReEig($\varepsilon$)}};
\draw[arrow] (bimap1) -- (re1);

\node[draw=black, thick, rounded corners, inner sep=0.2cm, 
      fit=(bimap1)(re1), label=above:{\textbf{BiRe block} 1}] (block1) {};

\node[right of=re1, node distance=1.8cm] (dots) {$\cdots$};
\draw[arrow] (re1) -- (dots);

\node[layer, right of=dots, node distance=1.8cm] (bimap2) {\textbf{BiMap}\\$\MAT{W}_L$};
\draw[arrow] (dots) -- (bimap2);
\node[layer, right of=bimap2] (re2) {\textbf{ReEig($\varepsilon$)}};
\draw[arrow] (bimap2) -- (re2);

\node[draw=black, thick, rounded corners, inner sep=0.2cm, 
      fit=(bimap2)(re2), label=above:{\textbf{BiRe block} $L$}] (block2) {};

\node[layer, right of=re2] (log) {\textbf{LogEig}};
\draw[arrow] (re2) -- (log);

\node[layer, right of=log] (fc) {\textbf{softmax}\\$\VEC{\xi},\VEC{\beta}$};
\draw[arrow] (log) -- (fc) node[midway,above]{\scriptsize vech};

\node[layer, fill=white, right of=fc] (output) {\textbf{Output}\\ $\hat{y}$};
\draw[arrow] (fc) -- (output);

\end{tikzpicture}
}
\caption{General SPDnet architecture~\cite{huang2017riemannian}: a chain of $L$ stacked \textbf{BiRe} (\textbf{BiMap} + \textbf{ReEig}) blocks, followed by \textbf{LogEig} mapping to Euclidean space and a fully connected softmax classifier. Learnable parameters are Stiefel weights $\MAT{W}_1,\dots, \MAT{W}_L$ and Euclidean weights $\VEC{\xi}$, and bias $\VEC{\beta}$.}
\label{fig:SPDnet-architecture}
\end{figure*}
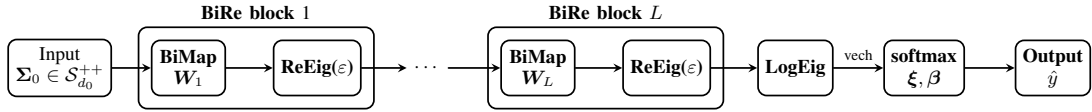

SPDnet~\cite{huang2017riemannian} is a deep learning architecture (see Figure~\ref{fig:SPDnet-architecture}) operating on SPD (covariance) matrices in a SPD-preserving way while reducing dimension thanks to \textbf{BiRe} blocks.
Such blocks combine a \textbf{BiMap} layer, performing bilinear projection for dimensionality reduction: \(
\MAT{\bar{\Sigma}}_\ell = \MAT{W}_\ell^\top\MAT{\Sigma}_{\ell-1}\MAT{W}_{\ell}, \, \MAT{W}_\ell\in \St_{d_{\ell-1},d_\ell},
\)
and a \textbf{ReEig} layer inducing non-linearity by eigenvalues rectification:
\(
\MAT{\Sigma}_\ell = \MAT{U}\max(\varepsilon \MAT{I}_{d_\ell},\MAT{\Lambda})\MAT{U}^\top, \, \MAT{\bar{\Sigma}}_{\ell}=\MAT{U}\MAT{\Lambda}\MAT{U}^\top.
\)
After $L$ \textbf{BiRe} blocks, the representation $\MAT{\Sigma}_{L-1}$ is mapped from the SPD manifold to the Euclidean space through the \textbf{LogEig} layer
\(
\MAT{\Sigma}_L = \MAT{U}\log(\MAT{\Lambda})\MAT{U}^\top, \, \MAT{\Sigma}_{L-1}=\MAT{U}\MAT{\Lambda}\MAT{U}^\top
\)
and finally fed to a softmax classifier via half-vectorization,
\(
\hat{y}=\text{softmax}(\VEC{\xi}\mathrm{vech}(\MAT{\Sigma}_L)+\VEC{\beta}), \, 
\VEC{\xi}\in\mathbb{R}^{K\times\nicefrac{d_L(d_L+1)}{2}},\,\VEC{\beta}\in\mathbb{R}^K,
\)
where $K$ is the number of classes.
The learnable parameters are
\(
\theta = (\MAT{W}_1,\dots,\MAT{W}_{L},\VEC{\xi},\VEC{\beta}).
\)

\section{Federated Learning with SPDnet Architecture}

In this section, we present our main contribution: FedSPDnet, a federated learning framework for the SPDnet architecture. The difficulty lies in the server-side aggregation, constrained by the Stiefel geometry of the \textbf{BiMap} weights, whose Euclidean average is generally not orthogonal and thus leaves the manifold. The Riemannian mean that would respect this geometry is intractable or too costly to compute at every round and for every layer~\cite{boumal2023introduction}, so existing manifold FL methods~\cite{li2022federated,huang2026federated,huang2024riemannian,zhang2024nonconvex} do not transfer to SPDnet. We instead introduce, in Section~\ref{subsec:orthoagg}, two efficient Stiefel aggregation strategies, ProjAvg and RLAvg. FedSPDnet is summarized in Algorithm~\ref{alg:fedSPDnet-avg}.

\subsection{Proposed model}

Let $\mathcal{C}=\{1,\dots,N\}$ denote the set of all clients.
The client $i$ has the training dataset
\begin{equation}
    \mathcal{D}^{(i)} = \left\{(\MAT{\Sigma}_j^{(i)}, y_j^{(i)})\in\mathcal{S}^{++}_{d_0}\times\llbracket1,K\rrbracket\right\}_{j=1}^{|\mathcal{D}^{(i)}|},
\end{equation}
where $y_j^{(i)}$ is the class label associated with the covariance $\MAT{\Sigma}_j^{(i)}$ and $|\mathcal{D}^{(i)}|$ is the cardinal of the dataset $\mathcal{D}^{(i)}$.
Given SPDnet learnable parameters $\theta=(\MAT{W}_1,\dots,\MAT{W}_L,\VEC{\xi},\VEC{\beta})$, with, $\forall\ell\in\llbracket1,L\rrbracket$, BiMap weights $\MAT{W}_{\ell}\in\St_{d_{\ell-1},d_\ell}$, and softmax parameters $\VEC{\xi}\in\R^{K\times\nicefrac{d_L(d_L+1)}{2}}$ and $\VEC{\beta}\in\R^K$, the empirical risk based on cross-entropy ~\cite{goodfellow2016deep} on client $i$ is
\begin{equation}
F_i(\theta) = \frac{-1}{|\mathcal{D}^{(i)}|} \sum_{j=1}^{|\mathcal{D}^{(i)}|}\sum_{k=1}^K \delta_{y_j^{(i)}=k} \log\left([f_{\theta}(\MAT{\Sigma}_j^{(i)})]_k\right),
\end{equation}
where $\delta_{y_j^{(i)}=k}$ is the Kronecker delta and $f_{\theta}(\MAT{\Sigma}_j^{(i)})$ denotes the forward pass of $\MAT{\Sigma}_j^{(i)}$ through SPDnet.

On the server, the objective is to minimize the global empirical risk over all clients, \textit{i.e.}, solve
\begin{equation}
    \argmin_{\theta} \quad \frac{1}{N}\sum_{i=1}^N F_i(\theta).
\end{equation}
To find a solution collaboratively, $T$ client-server communication rounds are performed.
At each round $t$, a subset $\mathcal{S}_t\subset\mathcal{C}$ of $M$ clients is selected uniformly at random.

The first step of the round $t$ consists in optimizing every local SPDnet model on the selected clients.
For each client $i\in\mathcal{S}_t$, the global parameters $\theta_t=(\MAT{W}_{1,t},\dots,\MAT{W}_{L,t},\VEC{\xi}_t,\VEC{\beta}_t)$ are set to the local SPDnet architecture.
Then, client $i$ runs $E_i$ local epochs of standard SPDnet training with any manifold-aware optimizer, and returns updated parameters $\theta_t^{(i)}$ to the server.

From there, the second step, which is the key issue in this work, is to aggregate the set of parameters $\{\theta^{(i)}_t\}_{i\in\mathcal{S}_t}$ from every selected client to obtain the new global parameters $\theta_{t+1}$.
For the Euclidean softmax parameters, the usual averaging technique from~\cite{mcmahan2017communication} can be employed, \textit{i.e.},
\begin{equation}
    \VEC{\xi}_{t+1} = \frac{1}{|\mathcal{S}_t|}\sum_{i\in \mathcal{S}_t}\VEC{\xi}_t^{(i)},
    \quad
    \VEC{\beta}_{t+1} = \frac{1}{|\mathcal{S}_t|}\sum_{i\in S_t}\VEC{\beta}_{t}^{(i)}.
\label{eq:eucl_FedAvg}
\end{equation}
However, this standard procedure cannot be applied to the Stiefel weights.
This issue is overcome in the next subsection.

\subsection{Orthogonality-preserving aggregation}
\label{subsec:orthoagg}

In this subsection, we propose two strategies to aggregate weights $\{\MAT{W}_{\ell,t}^{(i)}\}_{i\in\mathcal{S}_t}$ in the Stiefel manifold $\St_{d_{\ell-1},d_\ell}$.
The first one is a straightforward adaptation of the usual federated learning averaging method~\cite{mcmahan2017communication}.
It relies on computing a suitable mean on the Stiefel manifold. As discussed in Section~\ref{ssec:riem-backgd}, the Riemannian mean on Stiefel has no closed form and is costly to approximate.
Following~\cite{bouchard2025beyond}, we instead project the arithmetic mean of the local weights onto the manifold.
Concretely, given weights $\{\MAT{W}_{\ell,t}^{(i)}\}_{i\in\mathcal{S}_t}$, the aggregated weight reads
\begin{equation}
    \MAT{W}_{\ell,t+1} = \mathcal{P}\left( \frac{1}{|\mathcal{S}_t|} \sum_{i\in\mathcal{S}_t} \MAT{W}_{\ell,t}^{(i)} \right),
    \label{eq:ProjAvg}
\end{equation}
where the projection $\mathcal{P}$ is defined in~\eqref{eq:st_retr}.
As explained in~\cite{bouchard2025beyond}, this projection is the one that best suits the geometry of the Stiefel manifold.
In the following, the corresponding FedSPDnet method is denoted ProjAvg.

Our second strategy is inspired by~\cite{huang2024riemannian}, where the authors avoid averaging on the manifold by noticing that the classical Euclidean federated average can be rewritten as the previous global parameter plus an average of the local empirical-risk gradients. Adapted to the Riemannian case, this yields an update requiring only a tangent-space average at the previous global iterate, which is then mapped back to the manifold through a retraction. While this avoids computing a mean on the manifold, the gradient-based rewriting ties the method to a specific Riemannian stochastic gradient descent as the local optimizer, which may not suit every client.
To overcome this issue, we adopt the same strategy with a different rewriting of the Euclidean average.
If the weights are Euclidean, we get the same update formula as in~\eqref{eq:eucl_FedAvg}, which can be rewritten as
\begin{equation}
    \MAT{W}_{\ell,t+1}
    = \frac{1}{|\mathcal{S}_t|} \sum_{i\in\mathcal{S}_t} \MAT{W}_{\ell,t}^{(i)}
    = \MAT{W}_{\ell,t} + \frac{1}{|\mathcal{S}_t|} \sum_{i\in\mathcal{S}_t} (\MAT{W}_{\ell,t}^{(i)} - \MAT{W}_{\ell,t}).
\end{equation}
One can then notice that we have
\begin{equation}
    \MAT{W}_{\ell,t+1} = \exp_{\MAT{W}_{\ell,t}}\left( 
        \frac{1}{|\mathcal{S}_t|} \sum_{i\in\mathcal{S}_t} \log_{\MAT{W}_{\ell,t}}(\MAT{W}_{\ell,t}^{(i)})
    \right),
\end{equation}
where $\exp_{\MAT{W}}(\MAT{X})=\MAT{W}+\MAT{X}$ and $\log_{\MAT{W}}(\MAT{\bar{W}})=\MAT{\bar{W}}-\MAT{W}$ denote the Euclidean exponential and logarithm mappings, respectively.
Unfortunately, we cannot apply this formula directly on the Stiefel manifold because the Riemannian logarithm has no closed form and the Riemannian exponential is numerically expensive and might not be advantageous~\cite{boumal2023introduction}.
Instead, as in~\cite{bouchard2025beyond}, we exploit some approximation of these tools.
The Riemannian exponential and logarithm can be replaced with a retraction $R$ and a lifting $L$, respectively.
For Stiefel, as discussed in~\cite{bouchard2025beyond}, a convenient choice is the polar retraction~\eqref{eq:st_retr} together with the lifting
\begin{equation}
    L_{\MAT{W}}(\MAT{\bar{W}}) = P_{\MAT{W}}(\MAT{\bar{W}}-\MAT{W}),
\end{equation}
where $P$ is defined in~\eqref{eq:st_tangent_proj}.
Finally, the aggregation formula is
\begin{equation}
    \MAT{W}_{\ell,t+1} = R_{\MAT{W}_{\ell,t}}\left( 
        \frac{1}{|\mathcal{S}_t|} \sum_{i\in\mathcal{S}_t} L_{\MAT{W}_{\ell,t}}(\MAT{W}_{\ell,t}^{(i)})
    \right).
    \label{eq:RLAvg}
\end{equation}
In the following, this corresponding aggregation method is denoted RLAvg.
The resulting algorithm is provided in Algorithm~\ref{alg:fedSPDnet-avg}.

\begin{algorithm}[t]
\caption{FedSPDnet}
\label{alg:fedSPDnet-avg}
\textbf{Input:} number of rounds $T$, number of clients per round $M$, number of local epochs $\{E_i\}_{i\in \mathcal{C}}$
\begin{algorithmic}[1]
\State Initialize randomly $\theta_0=(\MAT{W}_{1,0},\dots,\MAT{W}_{L,0},\VEC{\xi}_0,\VEC{\beta}_0)$
\For{$t=0,\ldots,T-1$}
    \State Randomly sample $S_t \subset \mathcal{C}$ with $|S_t|=M$
    \For{\textbf{each} client $i \in S_t$ \textbf{in parallel}}
        \State Initialize local SPDnet with $\theta_t$
        \State Train SPDnet over $E_i$ epochs to get $\theta_{t}^{(i)}$
        \State Send $\theta_{t}^{(i)}$ to the server
    \EndFor
    \State Get $\VEC{\xi}_{t+1}$ and $\VEC{\beta}_{t+1}$ with~\eqref{eq:eucl_FedAvg}
    \For{$\ell=1,\ldots,L$}
        \State Compute $\MAT{W}_{\ell,t+1}$ with either~\eqref{eq:ProjAvg} or~\eqref{eq:RLAvg}
    \EndFor
    \State $\theta_{t+1} = \big(\MAT{W}_{1,t+1},\dots,\MAT{W}_{L,t+1},\VEC{\xi}_{t+1},\VEC{\beta}_{t+1}\big)$
\EndFor
\end{algorithmic}
\textbf{Output:} $\theta_T$
\end{algorithm}


\section{Numerical experiments}
\label{sec:experiments}

We evaluate FedSPDnet on EEG-based motor imagery classification, where the spatial covariance matrix of a trial is a standard discriminative representation~\cite{barachant2012multiclass,li2017second} and multi-site data confidentiality naturally motivates federated learning.
Unlike traditional BCI pipelines~\cite{lotte2018review,carrara2025geometric} that rely on within-subject calibration, we target population-level models that generalize across subjects without per-user data, a regime motivated by clinical settings where recordings are distributed across sites and cannot be centralized.
As a Euclidean baseline, we consider EEGnet~\cite{lawhern2018eegnet}, a compact convolutional network operating on raw signals, federated with standard FedAvg~\cite{mcmahan2017communication}.

\subsection{Experimental setup}
\label{ssec:setup}
We use two motor imagery datasets from the MOABB benchmark~\cite{chevallier2024largest}:
\textbf{Weibo2014} (10 subjects, 60 channels, sampling frequency $f_s=200$\,Hz, 7 classes, ${\sim}80$ trials/class) and
\textbf{PhysionetMI} (106 subjects, 64 channels, sampling frequency $f_s=160$\,Hz, 4 classes, ${\sim}23$ trials/class).
Three PhysionetMI subjects were excluded due to incomplete recordings.
All signals are band-pass filtered in $[8, 32]$\,Hz.

\begin{figure*}[t]
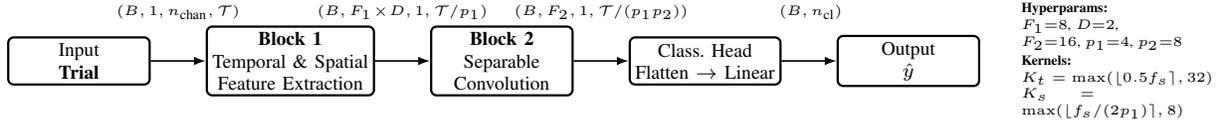

   \centering
   \resizebox{0.9\linewidth}{!}{\EEGNetGeneralFig}
   \caption{EEGnet overall pipeline.}
   \label{fig:eegnet-diagrams}
\end{figure*}

EEGnet~\cite{lawhern2018eegnet} operates on $z$-score normalized raw trials $X_j \in \mathbb{R}^{n_{\mathrm{chan}} \times \mathcal{T}}$, with $n_{\mathrm{chan}}$ channels and $\mathcal{T}$ time samples, through temporal, depthwise spatial, and separable convolutions followed by a classification head (Figure~\ref{fig:eegnet-diagrams}).

SPDnet~\cite{huang2017riemannian} operates on the sample covariance matrix (SCM) of each centered trial
\(
\Sigma_j = \frac{1}{\mathcal{T}-1}\,\bar{X}_j \bar{X}_j^\top \in \mathcal{S}_{n_{\mathrm{chan}}}^{++},
\)
so the input dimension is $d_0 = n_{\mathrm{chan}}$.
A single BiRe block is used, with hidden dimension~$d$ and ReEig threshold~$\varepsilon$ chosen per dataset.
Prior spectral analysis of the sample covariances sets~$\varepsilon$ to balance stability against discriminative information ($\varepsilon = 10^{-1}$ for Weibo2014, $10^{-2}$ for PhysionetMI), and a sweep over $d$ retains the smallest value maximising validation F1 ($d = 22$ for Weibo2014, $d = 18$ for PhysionetMI), yielding models more compact than EEGnet.

Both architectures are trained with the cross-entropy loss using Adam~\cite{kingma2014adam}. For SPDnet, the Stiefel weights $\MAT{W}_\ell$ are optimized via tangent-space local trivialization~\cite{ablin2024infeasible} while the softmax parameters $\VEC{\xi},\VEC{\beta}$ are updated in the standard Euclidean way. We use initial learning rate $10^{-3}$, batch size~64, and ReduceLROnPlateau scheduling (patience~20, factor~0.5).
All results are reported over 10 independent runs.

\subsection{Centralized baselines}
\label{ssec:centralized}

As an upper bound on federated performance, we pool all subjects and perform a stratified split into training (75\%), validation (10\%), and test (15\%) sets.
Training runs for at most 300 epochs with early stopping (patience~75), retaining the best validation checkpoint~\cite{prechelt1998early}.
Results (Table~\ref{tab:consolidated-results}) show comparable centralized performance: SPDnet slightly outperforms EEGnet on Weibo2014 while EEGnet leads on PhysionetMI.

\subsection{Federated learning experiments}
\label{ssec:fl}
We simulate a multi-institutional setting by partitioning subjects into clients, with two subjects per client to model small data silos with non-identical subject distributions, yielding $N=5$ clients for Weibo2014 and $N=53$ for PhysionetMI.
Each client applies the same 75/10/15 stratified split to its local data.
Training follows Algorithm~\ref{alg:fedSPDnet-avg} with $T=150$ rounds and $E_i=2$ local epochs for all clients $i$, matching the centralized budget of 300 total epochs.
We consider both full participation (all clients every round) and half participation ($M=\lfloor N/2 \rfloor$ clients).

For FedEEGnet, we apply FedAvg excluding batch normalization statistics from aggregation~\cite{li2021fedbn}.

\subsection{Results and analysis}
\label{ssec:results}

\input{convergence_figure}
\begin{table}[t]
\centering
\caption{Final test F1 (\%) for centralized and federated approaches. Mean $\pm$ std over 10 seeds; best per row in \textbf{bold}.}
\label{tab:consolidated-results}
\scriptsize
\setlength{\tabcolsep}{3pt}
\begin{tabular}{@{}lccc@{}}
\toprule
\textbf{Dataset} & & \shortstack{\textbf{EEGnet}\\\scriptsize\emph{(FedAvg)}} & \shortstack{\textbf{SPDnet}\\\scriptsize\emph{(FedSPDnet)}} \\
\midrule
\multirow{4}{*}{Weibo2014}
  & \emph{\# params} & \emph{5\,303} & \emph{4\,715} \\
  & Centr. & $50.7 \pm 1.5$ & $\mathbf{51.7 \pm 0.8}$ \\
  & Full & $39.9 \pm 2.4$ & $\mathbf{43.3 \pm 1.0}$ \\
  & Half & $36.4 \pm 4.1$ & $\mathbf{41.2 \pm 0.8}$ \\
\midrule
\multirow{4}{*}{PhysionetMI}
  & \emph{\# params} & \emph{3\,284} & \emph{2\,452} \\
  & Centr. & $\mathbf{50.8 \pm 0.7}$ & $43.1 \pm 0.5$ \\
  & Full & $38.3 \pm 1.8$ & $\mathbf{39.5 \pm 0.6}$ \\
  & Half & $38.0 \pm 2.7$ & $\mathbf{39.5 \pm 0.9}$ \\
\bottomrule
\end{tabular}
\end{table}

\paragraph{Aggregation scheme equivalence}

ProjAvg and RLAvg share the same per-round complexity of $\mathcal{O}(Mn_{\text{chan}}d + n_{\text{chan}}d^2)$.
As shown in Figures~\ref{fig:conv-weibo} and~\ref{fig:conv-physionet}, their validation F1 curves overlap throughout all communication rounds on both datasets, under both full and partial participation. This empirical agreement is expected: when the local weights stay close to the global iterate, the lifting--retraction update~\eqref{eq:RLAvg} reduces to the polar projection of the arithmetic mean~\eqref{eq:ProjAvg} to first order, so the two aggregations coincide up to higher-order terms.
Since ProjAvg has slightly lower constant factors and does not require storing the previous global iterate, it is adopted as the default in Table~\ref{tab:consolidated-results}.
\paragraph{Federated performance}
Under full participation, FedSPDnet stays within $8.4$ and $3.6$ F1 points of its centralized scores on Weibo2014 ($43.3$ vs.\ $51.7$) and PhysionetMI ($39.5$ vs.\ $43.1$), whereas FedEEGnet loses $10.8$ and $12.5$ points (Table~\ref{tab:consolidated-results}). FedSPDnet thus retains more of its centralized accuracy while using fewer parameters. This advantage holds even on PhysionetMI, where SPDnet trails EEGnet centrally yet retains a larger fraction of its centralized accuracy under federation, which we tentatively attribute to the Stiefel constraint.
FedSPDnet also converges faster, reaching its plateau in fewer communication rounds while FedEEGnet continues to oscillate or improve slowly (Figures~\ref{fig:conv-weibo} and~\ref{fig:conv-physionet}), and exhibits lower variance across seeds.
\section{Conclusion}
We proposed FedSPDnet, a federated learning framework for SPDnet with two optimizer-agnostic and geometry-preserving aggregation strategies (ProjAvg and RLAvg) that rely only on standard linear algebra.
Both are empirically equivalent but ProjAvg is recommended for its simpler, storage-free implementation.
On two EEG motor imagery benchmarks, FedSPDnet retains more of its centralized accuracy than federated EEGnet, is more robust to partial participation than federated EEGnet and converges faster while communicating fewer parameters per round. We emphasize that keeping data on each client does not by itself guarantee privacy: the communicated weights remain vulnerable to attacks. Endowing FedSPDnet with differential privacy~\cite{huang2026federated}, where the manifold constraint raises additional sensitivity and calibration questions, is left for future work.

\bibliographystyle{IEEEtran}
\bibliography{references}

\end{document}

%% file: eegnet_figures.tex
\usetikzlibrary{arrows.meta, calc, positioning, fit, decorations.pathreplacing, shapes.geometric, backgrounds}

\tikzset{
  eegbox/.style={
    draw, thick, rounded corners=2pt, align=center,
    font=\scriptsize, minimum height=6mm, minimum width=18mm,
    fill=white, text=black, inner sep=2pt
  },
  eegsmall/.style={
    draw, thick, rounded corners=2pt, align=center,
    font=\tiny, minimum height=5mm, inner sep=2pt,
    fill=white, text=black
  },
  eegarrow/.style={-{Latex[length=1.5mm]}, semithick},
  note/.style={font=\tiny, align=left, text=black},
  eegshapelabel/.style={midway, above, font=\tiny, yshift=4mm},
  eeggroup/.style={
    draw, thick, rounded corners=4pt, inner sep=4pt, fill=white
  }
}

\newcommand{\EEGNetGeneralFig}{%
\begin{tikzpicture}[node distance=5mm and 7mm]
\node[eegbox] (inp) {Input\\\textbf{Trial}};
\node[eegbox, right=of inp] (block1) {\textbf{Block 1}\\Temporal \& Spatial\\Feature Extraction};
\node[eegbox, right=of block1] (block2) {\textbf{Block 2}\\Separable\\Convolution};
\node[eegbox, right=of block2] (head) {Class.\ Head\\ Flatten $\to$ Linear};
\node[eegbox, right=of head] (out) {Output\\$\hat{y}$};

\draw[eegarrow] (inp) -- node[eegshapelabel]
  {$(B,1,n_{\text{chan}},\mathcal{T})$} (block1);
\draw[eegarrow] (block1) -- node[eegshapelabel]
  {$(B,F_1\!\times\! D,1,\mathcal{T}/p_1)$} (block2);
\draw[eegarrow] (block2) -- node[eegshapelabel]
  {$(B,F_2,1,\mathcal{T}/(p_1p_2))$} (head);
\draw[eegarrow] (head) -- node[eegshapelabel]
  {$\,(B,n_{\text{cl}})$} (out);

\node[note, right=4mm of out, text width=26mm] {%
\textbf{Hyperparams:}\\
$F_1{=}8$, $D{=}2$,\\
$F_2{=}16$,
$p_1{=}4$, $p_2{=}8$\\[1pt]
\textbf{Kernels:}\\
$K_t=\max(\lfloor 0.5 f_s \rceil,32)$\\
$K_s=\max(\lfloor f_s/(2p_1) \rceil,8)$
};
\end{tikzpicture}%
}

%% file: convergence_figure.tex

\definecolor{eegnetcolor}{HTML}{E69F00}
\definecolor{projavgcolor}{HTML}{0072B2}
\definecolor{rlavgcolor}{HTML}{009E73}

\begin{figure}[t]
\centering
\begin{tikzpicture}
\begin{axis}[
    width=0.85\columnwidth,
    height=0.51\columnwidth,
    xlabel={Round},
    ylabel={Test F1 (\%)},
    label style={font=\scriptsize},
    tick label style={font=\tiny},
    xmin=0, xmax=150,
    ymin=5, ymax=48,
    grid=major,
    grid style={dashed, gray!30},
    axis lines=box,
    axis line style={black},
    tick style={black},
    legend columns=3,
    legend style={
        at={(0.5,1.02)},
        anchor=south,
        font=\tiny,
        draw=none,
        fill=none,
        column sep=3pt,
    },
]
\addplot[eegnetcolor, draw=none, name path=w_eeg100_u, forget plot] coordinates {
(1,12.21)(4,19.73)(7,22.52)(10,26.31)(13,27.61)(16,29.55)(19,31.08)(22,32.43)(25,33.86)(28,34.23)(31,33.84)(34,35.22)(37,35.95)(40,36.16)(43,36.42)(46,36.65)(49,36.53)(52,36.85)(55,38.42)(58,37.60)(61,38.23)(64,39.00)(67,39.08)(70,38.45)(73,39.29)(77,40.57)(80,39.50)(83,40.20)(86,39.99)(89,41.13)(92,39.83)(95,39.90)(98,41.30)(101,40.93)(104,40.20)(107,41.88)(110,40.71)(113,41.69)(116,40.86)(119,40.80)(122,41.75)(125,39.45)(128,42.18)(131,41.68)(134,41.28)(137,41.50)(140,41.79)(143,42.40)(146,42.14)(150,42.25)
};
\addplot[eegnetcolor, draw=none, name path=w_eeg100_l, forget plot] coordinates {
(1,8.09)(4,16.15)(7,19.84)(10,22.53)(13,25.20)(16,26.92)(19,27.69)(22,29.54)(25,29.81)(28,29.98)(31,30.69)(34,30.67)(37,31.67)(40,32.01)(43,32.51)(46,33.22)(49,32.36)(52,33.02)(55,33.66)(58,33.86)(61,33.88)(64,34.01)(67,34.79)(70,34.73)(73,35.23)(77,36.02)(80,34.74)(83,35.64)(86,33.43)(89,35.19)(92,35.80)(95,35.21)(98,36.23)(101,36.72)(104,35.54)(107,37.44)(110,36.43)(113,38.45)(116,36.98)(119,37.40)(122,37.26)(125,35.83)(128,35.46)(131,36.76)(134,38.26)(137,37.74)(140,37.99)(143,36.11)(146,37.34)(150,37.55)
};
\addplot[eegnetcolor, fill opacity=0.1, forget plot] fill between[of=w_eeg100_u and w_eeg100_l];
\addplot[projavgcolor, draw=none, name path=w_proj100_u, forget plot] coordinates {
(1,19.06)(4,26.80)(7,29.12)(10,30.42)(13,31.71)(16,33.19)(19,34.29)(22,34.65)(25,35.43)(28,36.24)(31,36.92)(34,37.98)(37,38.51)(40,39.00)(43,39.81)(46,40.30)(49,40.52)(52,40.98)(55,41.58)(58,41.92)(61,42.26)(64,42.37)(67,42.10)(70,43.08)(73,43.23)(77,43.21)(80,43.70)(83,43.89)(86,43.30)(89,43.39)(92,43.53)(95,44.20)(98,43.99)(101,44.14)(104,44.20)(107,43.26)(110,44.36)(113,44.19)(116,43.89)(119,44.21)(122,43.98)(125,44.06)(128,44.17)(131,44.59)(134,44.60)(137,44.57)(140,44.23)(143,44.63)(146,44.49)(150,44.33)
};
\addplot[projavgcolor, draw=none, name path=w_proj100_l, forget plot] coordinates {
(1,13.36)(4,22.88)(7,25.75)(10,27.72)(13,27.99)(16,30.37)(19,30.88)(22,31.72)(25,32.44)(28,33.04)(31,33.97)(34,34.61)(37,35.60)(40,36.23)(43,37.76)(46,37.08)(49,38.25)(52,37.92)(55,38.18)(58,38.95)(61,39.15)(64,39.99)(67,39.83)(70,39.68)(73,40.42)(77,40.69)(80,40.66)(83,41.24)(86,41.16)(89,40.61)(92,41.12)(95,41.29)(98,41.23)(101,41.36)(104,41.61)(107,41.68)(110,42.08)(113,41.47)(116,41.81)(119,41.88)(122,41.79)(125,41.88)(128,41.79)(131,42.21)(134,42.23)(137,41.92)(140,42.56)(143,42.07)(146,42.15)(150,42.29)
};
\addplot[projavgcolor, fill opacity=0.1, forget plot] fill between[of=w_proj100_u and w_proj100_l];
\addplot[rlavgcolor, draw=none, name path=w_rl100_u, forget plot] coordinates {
(1,19.06)(4,26.83)(7,29.11)(10,30.40)(13,31.72)(16,33.20)(19,34.24)(22,34.70)(25,35.51)(28,36.27)(31,36.88)(34,37.93)(37,38.47)(40,38.94)(43,39.76)(46,40.31)(49,40.56)(52,40.99)(55,41.65)(58,41.88)(61,42.19)(64,42.42)(67,42.11)(70,43.05)(73,43.21)(77,43.10)(80,43.71)(83,43.86)(86,43.26)(89,43.32)(92,43.61)(95,44.21)(98,43.93)(101,44.19)(104,44.29)(107,43.31)(110,44.35)(113,44.17)(116,43.84)(119,44.26)(122,43.89)(125,44.00)(128,44.32)(131,44.74)(134,44.61)(137,44.49)(140,44.28)(143,44.58)(146,44.62)(150,44.41)
};
\addplot[rlavgcolor, draw=none, name path=w_rl100_l, forget plot] coordinates {
(1,13.36)(4,22.90)(7,25.76)(10,27.74)(13,27.99)(16,30.41)(19,30.91)(22,31.67)(25,32.41)(28,33.07)(31,33.95)(34,34.62)(37,35.57)(40,36.28)(43,37.79)(46,37.06)(49,38.36)(52,37.93)(55,38.21)(58,38.94)(61,39.09)(64,39.97)(67,39.83)(70,39.65)(73,40.43)(77,40.66)(80,40.66)(83,41.26)(86,41.10)(89,40.60)(92,41.09)(95,41.21)(98,41.21)(101,41.36)(104,41.51)(107,41.66)(110,42.10)(113,41.51)(116,41.76)(119,41.90)(122,41.86)(125,41.82)(128,41.75)(131,42.24)(134,42.18)(137,41.98)(140,42.60)(143,42.07)(146,42.14)(150,42.23)
};
\addplot[rlavgcolor, fill opacity=0.1, forget plot] fill between[of=w_rl100_u and w_rl100_l];
\addplot[eegnetcolor, draw=none, name path=w_eeg50_u, forget plot] coordinates {
(1,10.18)(4,18.11)(7,21.05)(10,24.37)(13,25.82)(16,29.75)(19,30.17)(22,29.65)(25,31.28)(28,31.17)(31,31.48)(34,34.59)(37,31.43)(40,34.92)(43,31.58)(46,32.96)(49,35.23)(52,34.98)(55,35.96)(58,33.55)(61,36.25)(64,37.11)(67,35.04)(70,34.64)(73,37.48)(77,37.17)(80,35.63)(83,37.85)(86,39.06)(89,39.60)(92,39.40)(95,39.62)(98,36.24)(101,37.85)(104,38.82)(107,38.88)(110,39.64)(113,41.55)(116,38.69)(119,38.79)(122,37.07)(125,38.15)(128,39.07)(131,37.48)(134,40.96)(137,40.75)(140,38.86)(143,38.82)(146,38.86)(150,40.48)
};
\addplot[eegnetcolor, draw=none, name path=w_eeg50_l, forget plot] coordinates {
(1,6.87)(4,14.00)(7,17.11)(10,20.37)(13,21.38)(16,24.08)(19,25.94)(22,23.80)(25,27.28)(28,27.09)(31,25.45)(34,29.22)(37,24.22)(40,23.05)(43,29.43)(46,28.57)(49,30.53)(52,28.79)(55,31.27)(58,30.31)(61,29.71)(64,31.54)(67,29.89)(70,31.86)(73,28.58)(77,26.36)(80,24.17)(83,28.50)(86,30.32)(89,29.32)(92,29.43)(95,25.45)(98,25.90)(101,30.93)(104,27.28)(107,33.57)(110,31.30)(113,26.77)(116,24.97)(119,26.95)(122,26.14)(125,25.01)(128,30.28)(131,27.04)(134,30.64)(137,28.02)(140,33.20)(143,28.16)(146,31.57)(150,32.38)
};
\addplot[eegnetcolor, fill opacity=0.08, forget plot] fill between[of=w_eeg50_u and w_eeg50_l];
\addplot[projavgcolor, draw=none, name path=w_proj50_u, forget plot] coordinates {
(1,15.36)(4,24.00)(7,27.50)(10,28.38)(13,30.61)(16,31.42)(19,33.49)(22,33.31)(25,35.50)(28,34.76)(31,35.82)(34,37.37)(37,37.61)(40,38.36)(43,38.26)(46,38.28)(49,39.00)(52,38.24)(55,39.85)(58,40.19)(61,40.93)(64,40.95)(67,40.27)(70,40.94)(73,41.17)(77,40.67)(80,42.62)(83,41.69)(86,41.66)(89,41.78)(92,41.86)(95,42.01)(98,41.86)(101,43.07)(104,42.47)(107,42.30)(110,42.85)(113,43.46)(116,42.82)(119,43.10)(122,42.80)(125,42.96)(128,42.48)(131,43.15)(134,43.73)(137,43.44)(140,43.38)(143,43.61)(146,44.92)(150,42.00)
};
\addplot[projavgcolor, draw=none, name path=w_proj50_l, forget plot] coordinates {
(1,11.09)(4,20.07)(7,23.43)(10,26.26)(13,27.56)(16,28.51)(19,29.67)(22,29.38)(25,30.66)(28,31.74)(31,32.51)(34,32.87)(37,33.97)(40,36.05)(43,34.93)(46,35.19)(49,35.97)(52,35.97)(55,35.75)(58,36.97)(61,36.46)(64,36.86)(67,36.52)(70,37.96)(73,37.71)(77,37.87)(80,38.82)(83,37.66)(86,40.11)(89,38.79)(92,39.81)(95,39.27)(98,39.15)(101,38.50)(104,39.26)(107,39.31)(110,39.71)(113,39.53)(116,40.08)(119,40.77)(122,39.90)(125,40.57)(128,38.85)(131,40.65)(134,40.47)(137,40.97)(140,41.10)(143,40.27)(146,40.75)(150,40.33)
};
\addplot[projavgcolor, fill opacity=0.08, forget plot] fill between[of=w_proj50_u and w_proj50_l];
\addplot[rlavgcolor, draw=none, name path=w_rl50_u, forget plot] coordinates {
(1,15.38)(4,23.99)(7,27.42)(10,28.44)(13,30.58)(16,31.46)(19,33.39)(22,33.36)(25,35.48)(28,34.70)(31,35.82)(34,37.30)(37,37.54)(40,38.40)(43,38.21)(46,38.30)(49,39.05)(52,38.19)(55,39.94)(58,40.23)(61,40.71)(64,40.84)(67,40.30)(70,40.93)(73,41.08)(77,40.68)(80,42.57)(83,41.73)(86,41.63)(89,41.69)(92,41.69)(95,42.04)(98,41.89)(101,43.03)(104,42.50)(107,42.27)(110,42.79)(113,43.46)(116,42.74)(119,43.10)(122,43.00)(125,42.82)(128,42.38)(131,43.13)(134,43.69)(137,43.45)(140,43.36)(143,43.72)(146,44.84)(150,42.02)
};
\addplot[rlavgcolor, draw=none, name path=w_rl50_l, forget plot] coordinates {
(1,11.09)(4,20.05)(7,23.45)(10,26.21)(13,27.55)(16,28.51)(19,29.71)(22,29.43)(25,30.63)(28,31.76)(31,32.48)(34,32.95)(37,33.96)(40,36.04)(43,34.84)(46,35.13)(49,36.05)(52,36.02)(55,35.82)(58,36.92)(61,36.46)(64,36.85)(67,36.53)(70,37.98)(73,37.66)(77,37.76)(80,38.65)(83,37.67)(86,40.16)(89,38.80)(92,39.65)(95,39.34)(98,39.12)(101,38.69)(104,39.32)(107,39.32)(110,39.80)(113,39.41)(116,40.03)(119,40.80)(122,39.85)(125,40.60)(128,38.84)(131,40.55)(134,40.44)(137,40.88)(140,41.13)(143,40.28)(146,40.90)(150,40.35)
};
\addplot[rlavgcolor, fill opacity=0.08, forget plot] fill between[of=w_rl50_u and w_rl50_l];
\addplot[eegnetcolor, thick, solid] coordinates {
(1.0,10.15)(4.0,17.94)(7.0,21.18)(10.0,24.42)(13.0,26.40)(16.0,28.23)(19.0,29.39)(22.0,30.99)(25.0,31.84)(28.0,32.11)(31.0,32.26)(34.0,32.94)(37.0,33.81)(40.0,34.09)(43.0,34.47)(46.0,34.93)(49.0,34.45)(52.0,34.94)(55.0,36.04)(58.0,35.73)(61.0,36.05)(64.0,36.50)(67.0,36.93)(70.0,36.59)(73.0,37.26)(77.0,38.29)(80.0,37.12)(83.0,37.92)(86.0,36.71)(89.0,38.16)(92.0,37.81)(95.0,37.56)(98.0,38.77)(101.0,38.83)(104.0,37.87)(107.0,39.66)(110.0,38.57)(113.0,40.07)(116.0,38.92)(119.0,39.10)(122.0,39.50)(125.0,37.64)(128.0,38.82)(131.0,39.22)(134.0,39.77)(137.0,39.62)(140.0,39.89)(143.0,39.26)(146.0,39.74)(150.0,39.90)
};
\addlegendentry{FedEEGNet\,100\%}
\addplot[projavgcolor, thick, solid, mark=square*, mark size=1pt, mark repeat=10, mark phase=0] coordinates {
(-0.5,16.21)(2.5,24.84)(5.5,27.43)(8.5,29.07)(11.5,29.85)(14.5,31.78)(17.5,32.59)(20.5,33.18)(23.5,33.93)(26.5,34.64)(29.5,35.44)(32.5,36.30)(35.5,37.05)(38.5,37.61)(41.5,38.79)(44.5,38.69)(47.5,39.38)(50.5,39.45)(53.5,39.88)(56.5,40.44)(59.5,40.70)(62.5,41.18)(65.5,40.96)(68.5,41.38)(71.5,41.83)(75.5,41.95)(78.5,42.18)(81.5,42.56)(84.5,42.23)(87.5,42.00)(90.5,42.33)(93.5,42.74)(96.5,42.61)(99.5,42.75)(102.5,42.90)(105.5,42.47)(108.5,43.22)(111.5,42.83)(114.5,42.85)(117.5,43.04)(120.5,42.88)(123.5,42.97)(126.5,42.98)(129.5,43.40)(132.5,43.42)(135.5,43.24)(138.5,43.40)(141.5,43.35)(144.5,43.32)(148.5,43.31)
};
\addlegendentry{ProjAvg\,100\%}
\addplot[rlavgcolor, thick, solid, mark=triangle*, mark size=1.2pt, mark repeat=10, mark phase=5] coordinates {
(2.5,16.21)(5.5,24.86)(8.5,27.44)(11.5,29.07)(14.5,29.85)(17.5,31.81)(20.5,32.57)(23.5,33.18)(26.5,33.96)(29.5,34.67)(32.5,35.41)(35.5,36.28)(38.5,37.02)(41.5,37.61)(44.5,38.77)(47.5,38.68)(50.5,39.46)(53.5,39.46)(56.5,39.93)(59.5,40.41)(62.5,40.64)(65.5,41.20)(68.5,40.97)(71.5,41.35)(74.5,41.82)(78.5,41.88)(81.5,42.19)(84.5,42.56)(87.5,42.18)(90.5,41.96)(93.5,42.35)(96.5,42.71)(99.5,42.57)(102.5,42.77)(105.5,42.90)(108.5,42.48)(111.5,43.22)(114.5,42.84)(117.5,42.80)(120.5,43.08)(123.5,42.87)(126.5,42.91)(129.5,43.03)(132.5,43.49)(135.5,43.39)(138.5,43.24)(141.5,43.44)(144.5,43.32)(147.5,43.38)(151.5,43.32)
};
\addlegendentry{RLAvg\,100\%}
\addplot[eegnetcolor, thick, dashed] coordinates {
(1.0,8.53)(4.0,16.06)(7.0,19.08)(10.0,22.37)(13.0,23.60)(16.0,26.92)(19.0,28.06)(22.0,26.72)(25.0,29.28)(28.0,29.13)(31.0,28.47)(34.0,31.91)(37.0,27.83)(40.0,28.98)(43.0,30.51)(46.0,30.77)(49.0,32.88)(52.0,31.88)(55.0,33.62)(58.0,31.93)(61.0,32.98)(64.0,34.33)(67.0,32.47)(70.0,33.25)(73.0,33.03)(77.0,31.77)(80.0,29.90)(83.0,33.18)(86.0,34.69)(89.0,34.46)(92.0,34.41)(95.0,32.54)(98.0,31.07)(101.0,34.39)(104.0,33.05)(107.0,36.22)(110.0,35.47)(113.0,34.16)(116.0,31.83)(119.0,32.87)(122.0,31.61)(125.0,31.58)(128.0,34.68)(131.0,32.26)(134.0,35.80)(137.0,34.38)(140.0,36.03)(143.0,33.49)(146.0,35.21)(150.0,36.43)
};
\addlegendentry{FedEEGNet\,50\%}
\addplot[projavgcolor, thick, dashed, mark=square, mark size=1pt, mark repeat=10, mark phase=0] coordinates {
(-0.5,13.22)(2.5,22.03)(5.5,25.47)(8.5,27.32)(11.5,29.09)(14.5,29.97)(17.5,31.58)(20.5,31.34)(23.5,33.08)(26.5,33.25)(29.5,34.16)(32.5,35.12)(35.5,35.79)(38.5,37.20)(41.5,36.60)(44.5,36.73)(47.5,37.49)(50.5,37.11)(53.5,37.80)(56.5,38.58)(59.5,38.69)(62.5,38.91)(65.5,38.39)(68.5,39.45)(71.5,39.44)(75.5,39.27)(78.5,40.72)(81.5,39.68)(84.5,40.88)(87.5,40.28)(90.5,40.83)(93.5,40.64)(96.5,40.51)(99.5,40.79)(102.5,40.87)(105.5,40.80)(108.5,41.28)(111.5,41.49)(114.5,41.45)(117.5,41.93)(120.5,41.35)(123.5,41.76)(126.5,40.67)(129.5,41.90)(132.5,42.10)(135.5,42.20)(138.5,42.24)(141.5,41.94)(144.5,42.84)(148.5,41.16)
};
\addlegendentry{ProjAvg\,50\%}
\addplot[rlavgcolor, thick, dashed, mark=triangle, mark size=1.2pt, mark repeat=10, mark phase=5] coordinates {
(2.5,13.24)(5.5,22.02)(8.5,25.44)(11.5,27.33)(14.5,29.07)(17.5,29.99)(20.5,31.55)(23.5,31.39)(26.5,33.05)(29.5,33.23)(32.5,34.15)(35.5,35.13)(38.5,35.75)(41.5,37.22)(44.5,36.52)(47.5,36.71)(50.5,37.55)(53.5,37.10)(56.5,37.88)(59.5,38.58)(62.5,38.58)(65.5,38.85)(68.5,38.41)(71.5,39.46)(74.5,39.37)(78.5,39.22)(81.5,40.61)(84.5,39.70)(87.5,40.89)(90.5,40.24)(93.5,40.67)(96.5,40.69)(99.5,40.51)(102.5,40.86)(105.5,40.91)(108.5,40.79)(111.5,41.30)(114.5,41.43)(117.5,41.38)(120.5,41.95)(123.5,41.43)(126.5,41.71)(129.5,40.61)(132.5,41.84)(135.5,42.06)(138.5,42.16)(141.5,42.25)(144.5,42.00)(147.5,42.87)(151.5,41.19)
};
\addlegendentry{RLAvg\,50\%}
\end{axis}
\end{tikzpicture}
\caption{Convergence on Weibo2014 under full (solid) and half (dashed) participation. Shaded bands: $\pm$1\,std over 10 seeds. ProjAvg/RLAvg curves slightly offset horizontally.}
\label{fig:conv-weibo}
\end{figure}
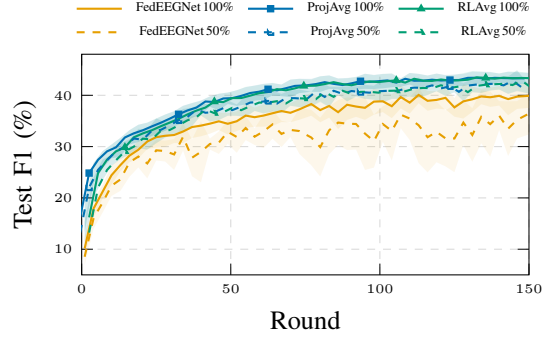

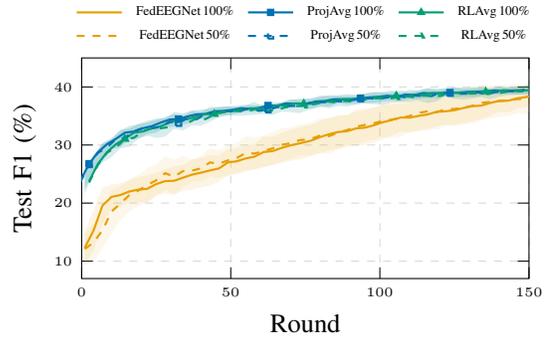
\begin{figure}[t]
\centering
\begin{tikzpicture}
\begin{axis}[
    width=0.85\columnwidth,
    height=0.51\columnwidth,
    xlabel={Round},
    ylabel={Test F1 (\%)},
    label style={font=\scriptsize},
    tick label style={font=\tiny},
    xmin=0, xmax=150,
    ymin=7, ymax=45,
    grid=major,
    grid style={dashed, gray!30},
    axis lines=box,
    axis line style={black},
    tick style={black},
    legend columns=3,
    legend style={
        at={(0.5,1.02)},
        anchor=south,
        font=\tiny,
        draw=none,
        fill=none,
        column sep=3pt,
    },
]
\addplot[eegnetcolor, draw=none, name path=p_eeg100_u, forget plot] coordinates {
(1,15.07)(4,18.86)(7,22.67)(10,23.21)(13,23.42)(16,23.87)(19,24.47)(22,24.79)(25,25.22)(28,25.87)(31,26.33)(34,26.58)(37,27.53)(40,27.78)(43,28.25)(46,28.21)(49,29.25)(52,29.45)(55,29.93)(58,30.19)(61,31.46)(64,31.36)(67,32.11)(70,32.60)(73,33.03)(77,33.50)(80,34.16)(83,34.70)(86,35.11)(89,35.09)(92,35.82)(95,35.99)(98,36.40)(101,36.92)(104,36.96)(107,37.13)(110,37.26)(113,37.66)(116,38.02)(119,38.18)(122,38.08)(125,38.55)(128,38.97)(131,39.30)(134,39.05)(137,39.44)(140,39.52)(143,39.65)(146,39.81)(150,40.14)
};
\addplot[eegnetcolor, draw=none, name path=p_eeg100_l, forget plot] coordinates {
(1,9.47)(4,11.74)(7,16.49)(10,18.92)(13,19.37)(16,20.13)(19,20.15)(22,20.12)(25,21.32)(28,21.70)(31,21.36)(34,22.18)(37,22.27)(40,22.70)(43,22.99)(46,23.72)(49,24.74)(52,24.95)(55,25.44)(58,25.82)(61,26.12)(64,26.54)(67,26.88)(70,27.27)(73,27.50)(77,28.02)(80,28.60)(83,29.03)(86,29.36)(89,29.68)(92,29.90)(95,30.00)(98,30.43)(101,30.93)(104,31.65)(107,32.34)(110,32.15)(113,32.38)(116,32.50)(119,33.26)(122,33.51)(125,33.42)(128,33.76)(131,34.09)(134,34.52)(137,35.03)(140,35.72)(143,35.60)(146,36.28)(150,36.54)
};
\addplot[eegnetcolor, fill opacity=0.1, forget plot] fill between[of=p_eeg100_u and p_eeg100_l];
\addplot[projavgcolor, draw=none, name path=p_proj100_u, forget plot] coordinates {
(1,25.15)(4,28.26)(7,30.22)(10,31.00)(13,32.50)(16,33.26)(19,33.54)(22,33.89)(25,34.34)(28,34.67)(31,35.64)(34,35.45)(37,35.95)(40,36.07)(43,36.23)(46,36.58)(49,36.66)(52,36.74)(55,36.93)(58,36.82)(61,37.03)(64,37.31)(67,37.33)(70,37.90)(73,37.77)(77,37.95)(80,38.29)(83,38.58)(86,38.55)(89,38.78)(92,38.65)(95,38.94)(98,39.10)(101,39.14)(104,39.38)(107,39.01)(110,39.51)(113,39.41)(116,39.82)(119,39.73)(122,39.63)(125,39.90)(128,39.71)(131,40.01)(134,39.94)(137,39.85)(140,40.10)(143,39.94)(146,40.08)(150,40.05)
};
\addplot[projavgcolor, draw=none, name path=p_proj100_l, forget plot] coordinates {
(1,22.40)(4,25.31)(7,27.30)(10,29.30)(13,29.63)(16,30.99)(19,31.10)(22,31.88)(25,32.14)(28,33.05)(31,33.15)(34,33.50)(37,33.88)(40,34.50)(43,34.48)(46,35.02)(49,35.09)(52,35.38)(55,35.75)(58,35.67)(61,35.77)(64,36.32)(67,36.24)(70,36.42)(73,36.48)(77,36.23)(80,36.72)(83,36.59)(86,37.19)(89,37.01)(92,37.11)(95,37.19)(98,37.31)(101,37.49)(104,37.62)(107,37.56)(110,37.85)(113,37.86)(116,38.08)(119,38.05)(122,38.27)(125,38.21)(128,38.36)(131,38.31)(134,38.50)(137,38.66)(140,38.74)(143,38.76)(146,38.71)(150,38.89)
};
\addplot[projavgcolor, fill opacity=0.1, forget plot] fill between[of=p_proj100_u and p_proj100_l];
\addplot[rlavgcolor, draw=none, name path=p_rl100_u, forget plot] coordinates {
(1,25.17)(4,28.27)(7,30.22)(10,30.98)(13,32.50)(16,33.26)(19,33.56)(22,33.89)(25,34.34)(28,34.67)(31,35.65)(34,35.45)(37,35.96)(40,36.07)(43,36.21)(46,36.59)(49,36.66)(52,36.72)(55,36.93)(58,36.82)(61,37.03)(64,37.31)(67,37.34)(70,37.91)(73,37.77)(77,37.94)(80,38.29)(83,38.58)(86,38.54)(89,38.78)(92,38.66)(95,38.95)(98,39.11)(101,39.14)(104,39.38)(107,38.99)(110,39.49)(113,39.40)(116,39.83)(119,39.73)(122,39.64)(125,39.90)(128,39.71)(131,39.98)(134,39.91)(137,39.85)(140,40.12)(143,39.93)(146,40.07)(150,40.05)
};
\addplot[rlavgcolor, draw=none, name path=p_rl100_l, forget plot] coordinates {
(1,22.39)(4,25.31)(7,27.30)(10,29.30)(13,29.63)(16,30.99)(19,31.10)(22,31.87)(25,32.14)(28,33.05)(31,33.16)(34,33.50)(37,33.89)(40,34.51)(43,34.48)(46,35.03)(49,35.09)(52,35.37)(55,35.75)(58,35.68)(61,35.77)(64,36.30)(67,36.24)(70,36.43)(73,36.48)(77,36.23)(80,36.72)(83,36.59)(86,37.19)(89,37.01)(92,37.11)(95,37.19)(98,37.29)(101,37.49)(104,37.63)(107,37.56)(110,37.84)(113,37.87)(116,38.08)(119,38.05)(122,38.25)(125,38.21)(128,38.36)(131,38.27)(134,38.49)(137,38.66)(140,38.77)(143,38.78)(146,38.70)(150,38.89)
};
\addplot[rlavgcolor, fill opacity=0.1, forget plot] fill between[of=p_rl100_u and p_rl100_l];
\addplot[eegnetcolor, draw=none, name path=p_eeg50_u, forget plot] coordinates {
(1,14.84)(4,16.52)(7,18.50)(10,22.34)(13,22.87)(16,24.19)(19,24.86)(22,26.02)(25,26.60)(28,28.05)(31,27.49)(34,28.28)(37,27.45)(40,28.16)(43,29.41)(46,29.14)(49,29.89)(52,30.37)(55,31.49)(58,31.25)(61,31.61)(64,32.26)(67,32.87)(70,33.03)(73,33.50)(77,33.85)(80,34.66)(83,34.54)(86,35.06)(89,35.23)(92,36.21)(95,36.35)(98,36.50)(101,36.99)(104,37.53)(107,37.94)(110,37.69)(113,38.48)(116,38.75)(119,38.65)(122,39.04)(125,39.20)(128,39.34)(131,39.79)(134,40.33)(137,40.28)(140,40.26)(143,40.44)(146,40.37)(150,40.67)
};
\addplot[eegnetcolor, draw=none, name path=p_eeg50_l, forget plot] coordinates {
(1,9.35)(4,9.86)(7,12.17)(10,14.74)(13,16.96)(16,18.57)(19,19.44)(22,20.18)(25,21.25)(28,22.24)(31,21.61)(34,22.75)(37,23.74)(40,23.77)(43,24.58)(46,24.39)(49,24.81)(52,25.18)(55,25.71)(58,25.82)(61,26.29)(64,26.58)(67,27.19)(70,27.55)(73,28.14)(77,28.28)(80,28.64)(83,28.53)(86,29.23)(89,29.94)(92,30.07)(95,30.59)(98,30.99)(101,31.35)(104,31.22)(107,31.38)(110,32.30)(113,32.58)(116,32.86)(119,32.62)(122,33.34)(125,33.22)(128,33.52)(131,33.70)(134,34.21)(137,34.57)(140,34.87)(143,34.92)(146,35.40)(150,35.33)
};
\addplot[eegnetcolor, fill opacity=0.08, forget plot] fill between[of=p_eeg50_u and p_eeg50_l];
\addplot[projavgcolor, draw=none, name path=p_proj50_u, forget plot] coordinates {
(1,25.49)(4,28.20)(7,30.39)(10,31.22)(13,32.63)(16,32.81)(19,33.68)(22,34.17)(25,34.58)(28,34.76)(31,35.31)(34,34.99)(37,35.88)(40,35.59)(43,36.65)(46,36.49)(49,36.65)(52,36.61)(55,36.79)(58,37.49)(61,36.41)(64,36.98)(67,37.25)(70,37.74)(73,38.01)(77,38.05)(80,38.56)(83,38.40)(86,38.43)(89,38.40)(92,38.48)(95,38.78)(98,39.09)(101,39.00)(104,39.03)(107,39.40)(110,39.10)(113,39.37)(116,39.68)(119,39.43)(122,39.83)(125,39.91)(128,39.89)(131,39.70)(134,39.92)(137,39.63)(140,39.85)(143,39.98)(146,39.84)(150,40.39)
};
\addplot[projavgcolor, draw=none, name path=p_proj50_l, forget plot] coordinates {
(1,21.59)(4,25.09)(7,26.97)(10,27.92)(13,29.43)(16,29.68)(19,31.08)(22,31.59)(25,31.23)(28,31.42)(31,32.79)(34,32.50)(37,33.41)(40,33.29)(43,34.00)(46,34.76)(49,35.11)(52,35.03)(55,35.10)(58,35.20)(61,35.37)(64,35.14)(67,35.77)(70,35.67)(73,35.47)(77,36.20)(80,36.51)(83,36.35)(86,36.72)(89,36.62)(92,36.91)(95,37.18)(98,37.01)(101,36.91)(104,37.36)(107,37.40)(110,37.39)(113,37.64)(116,37.22)(119,37.47)(122,38.42)(125,37.96)(128,37.78)(131,38.01)(134,38.25)(137,37.89)(140,38.32)(143,38.58)(146,38.67)(150,38.57)
};
\addplot[projavgcolor, fill opacity=0.08, forget plot] fill between[of=p_proj50_u and p_proj50_l];
\addplot[rlavgcolor, draw=none, name path=p_rl50_u, forget plot] coordinates {
(1,25.48)(4,28.18)(7,30.37)(10,31.22)(13,32.60)(16,32.81)(19,33.69)(22,34.17)(25,34.58)(28,34.75)(31,35.31)(34,35.02)(37,35.88)(40,35.59)(43,36.63)(46,36.49)(49,36.66)(52,36.60)(55,36.80)(58,37.49)(61,36.40)(64,36.95)(67,37.30)(70,37.74)(73,38.01)(77,38.06)(80,38.56)(83,38.41)(86,38.44)(89,38.41)(92,38.49)(95,38.76)(98,39.10)(101,39.00)(104,39.04)(107,39.41)(110,39.09)(113,39.37)(116,39.68)(119,39.45)(122,39.83)(125,39.93)(128,39.91)(131,39.74)(134,39.91)(137,39.64)(140,39.85)(143,39.99)(146,39.83)(150,40.39)
};
\addplot[rlavgcolor, draw=none, name path=p_rl50_l, forget plot] coordinates {
(1,21.58)(4,25.14)(7,26.98)(10,27.92)(13,29.44)(16,29.68)(19,31.08)(22,31.59)(25,31.23)(28,31.43)(31,32.79)(34,32.51)(37,33.41)(40,33.31)(43,34.00)(46,34.76)(49,35.13)(52,35.02)(55,35.10)(58,35.20)(61,35.37)(64,35.15)(67,35.74)(70,35.67)(73,35.47)(77,36.19)(80,36.52)(83,36.32)(86,36.73)(89,36.62)(92,36.89)(95,37.17)(98,37.02)(101,36.91)(104,37.33)(107,37.37)(110,37.37)(113,37.64)(116,37.22)(119,37.50)(122,38.41)(125,37.95)(128,37.80)(131,38.01)(134,38.24)(137,37.90)(140,38.32)(143,38.57)(146,38.66)(150,38.57)
};
\addplot[rlavgcolor, fill opacity=0.08, forget plot] fill between[of=p_rl50_u and p_rl50_l];
\addplot[eegnetcolor, thick, solid] coordinates {
(1.0,12.27)(4.0,15.30)(7.0,19.58)(10.0,21.06)(13.0,21.39)(16.0,22.00)(19.0,22.31)(22.0,22.46)(25.0,23.27)(28.0,23.78)(31.0,23.84)(34.0,24.38)(37.0,24.90)(40.0,25.24)(43.0,25.62)(46.0,25.96)(49.0,26.99)(52.0,27.20)(55.0,27.69)(58.0,28.00)(61.0,28.79)(64.0,28.95)(67.0,29.49)(70.0,29.94)(73.0,30.27)(77.0,30.76)(80.0,31.38)(83.0,31.87)(86.0,32.24)(89.0,32.38)(92.0,32.86)(95.0,32.99)(98.0,33.41)(101.0,33.92)(104.0,34.31)(107.0,34.74)(110.0,34.71)(113.0,35.02)(116.0,35.26)(119.0,35.72)(122.0,35.80)(125.0,35.99)(128.0,36.37)(131.0,36.69)(134.0,36.79)(137.0,37.24)(140.0,37.62)(143.0,37.63)(146.0,38.04)(150.0,38.34)
};
\addlegendentry{FedEEGNet\,100\%}
\addplot[projavgcolor, thick, solid, mark=square*, mark size=1pt, mark repeat=10, mark phase=0] coordinates {
(-0.5,23.77)(2.5,26.78)(5.5,28.76)(8.5,30.15)(11.5,31.07)(14.5,32.12)(17.5,32.32)(20.5,32.89)(23.5,33.24)(26.5,33.86)(29.5,34.39)(32.5,34.47)(35.5,34.92)(38.5,35.28)(41.5,35.36)(44.5,35.80)(47.5,35.88)(50.5,36.06)(53.5,36.34)(56.5,36.24)(59.5,36.40)(62.5,36.81)(65.5,36.78)(68.5,37.16)(71.5,37.13)(75.5,37.09)(78.5,37.50)(81.5,37.59)(84.5,37.87)(87.5,37.90)(90.5,37.88)(93.5,38.06)(96.5,38.21)(99.5,38.31)(102.5,38.50)(105.5,38.28)(108.5,38.68)(111.5,38.64)(114.5,38.95)(117.5,38.89)(120.5,38.95)(123.5,39.05)(126.5,39.03)(129.5,39.16)(132.5,39.22)(135.5,39.25)(138.5,39.42)(141.5,39.35)(144.5,39.39)(148.5,39.47)
};
\addlegendentry{ProjAvg\,100\%}
\addplot[rlavgcolor, thick, solid, mark=triangle*, mark size=1.2pt, mark repeat=10, mark phase=5] coordinates {
(2.5,23.78)(5.5,26.79)(8.5,28.76)(11.5,30.14)(14.5,31.07)(17.5,32.12)(20.5,32.33)(23.5,32.88)(26.5,33.24)(29.5,33.86)(32.5,34.41)(35.5,34.47)(38.5,34.92)(41.5,35.29)(44.5,35.35)(47.5,35.81)(50.5,35.88)(53.5,36.05)(56.5,36.34)(59.5,36.25)(62.5,36.40)(65.5,36.81)(68.5,36.79)(71.5,37.17)(74.5,37.13)(78.5,37.09)(81.5,37.50)(84.5,37.59)(87.5,37.86)(90.5,37.90)(93.5,37.89)(96.5,38.07)(99.5,38.20)(102.5,38.31)(105.5,38.51)(108.5,38.28)(111.5,38.67)(114.5,38.64)(117.5,38.96)(120.5,38.89)(123.5,38.94)(126.5,39.05)(129.5,39.03)(132.5,39.13)(135.5,39.20)(138.5,39.25)(141.5,39.44)(144.5,39.35)(147.5,39.39)(151.5,39.47)
};
\addlegendentry{RLAvg\,100\%}
\addplot[eegnetcolor, thick, dashed] coordinates {
(1.0,12.10)(4.0,13.19)(7.0,15.34)(10.0,18.54)(13.0,19.91)(16.0,21.38)(19.0,22.15)(22.0,23.10)(25.0,23.92)(28.0,25.15)(31.0,24.55)(34.0,25.51)(37.0,25.60)(40.0,25.96)(43.0,27.00)(46.0,26.77)(49.0,27.35)(52.0,27.78)(55.0,28.60)(58.0,28.54)(61.0,28.95)(64.0,29.42)(67.0,30.03)(70.0,30.29)(73.0,30.82)(77.0,31.06)(80.0,31.65)(83.0,31.53)(86.0,32.15)(89.0,32.58)(92.0,33.14)(95.0,33.47)(98.0,33.75)(101.0,34.17)(104.0,34.38)(107.0,34.66)(110.0,35.00)(113.0,35.53)(116.0,35.80)(119.0,35.64)(122.0,36.19)(125.0,36.21)(128.0,36.43)(131.0,36.75)(134.0,37.27)(137.0,37.42)(140.0,37.56)(143.0,37.68)(146.0,37.89)(150.0,38.00)
};
\addlegendentry{FedEEGNet\,50\%}
\addplot[projavgcolor, thick, dashed, mark=square, mark size=1pt, mark repeat=10, mark phase=0] coordinates {
(-0.5,23.54)(2.5,26.64)(5.5,28.68)(8.5,29.57)(11.5,31.03)(14.5,31.24)(17.5,32.38)(20.5,32.88)(23.5,32.90)(26.5,33.09)(29.5,34.05)(32.5,33.75)(35.5,34.65)(38.5,34.44)(41.5,35.32)(44.5,35.62)(47.5,35.88)(50.5,35.82)(53.5,35.94)(56.5,36.34)(59.5,35.89)(62.5,36.06)(65.5,36.51)(68.5,36.70)(71.5,36.74)(75.5,37.13)(78.5,37.54)(81.5,37.37)(84.5,37.57)(87.5,37.51)(90.5,37.69)(93.5,37.98)(96.5,38.05)(99.5,37.96)(102.5,38.19)(105.5,38.40)(108.5,38.24)(111.5,38.50)(114.5,38.45)(117.5,38.45)(120.5,39.12)(123.5,38.93)(126.5,38.84)(129.5,38.86)(132.5,39.09)(135.5,38.76)(138.5,39.09)(141.5,39.28)(144.5,39.26)(148.5,39.48)
};
\addlegendentry{ProjAvg\,50\%}
\addplot[rlavgcolor, thick, dashed, mark=triangle, mark size=1.2pt, mark repeat=10, mark phase=5] coordinates {
(2.5,23.53)(5.5,26.66)(8.5,28.68)(11.5,29.57)(14.5,31.02)(17.5,31.24)(20.5,32.38)(23.5,32.88)(26.5,32.90)(29.5,33.09)(32.5,34.05)(35.5,33.76)(38.5,34.65)(41.5,34.45)(44.5,35.32)(47.5,35.62)(50.5,35.89)(53.5,35.81)(56.5,35.95)(59.5,36.34)(62.5,35.88)(65.5,36.05)(68.5,36.52)(71.5,36.70)(74.5,36.74)(78.5,37.12)(81.5,37.54)(84.5,37.36)(87.5,37.59)(90.5,37.51)(93.5,37.69)(96.5,37.97)(99.5,38.06)(102.5,37.95)(105.5,38.18)(108.5,38.39)(111.5,38.23)(114.5,38.50)(117.5,38.45)(120.5,38.47)(123.5,39.12)(126.5,38.94)(129.5,38.86)(132.5,38.88)(135.5,39.07)(138.5,38.77)(141.5,39.09)(144.5,39.28)(147.5,39.24)(151.5,39.48)
};
\addlegendentry{RLAvg\,50\%}
\end{axis}
\end{tikzpicture}
\caption{Convergence on PhysionetMI  under full (solid) and half (dashed) participation. Shaded bands: $\pm$1\,std over 10 seeds. ProjAvg/RLAvg curves slightly offset horizontally.}
\label{fig:conv-physionet}
\end{figure}

%% file: references.bib
@article{huang2026federated,
  title={Federated learning on {Riemannian} manifolds with differential privacy},
  author={Huang, Zhenwei and Huang, Wen and Jawanpuria, Pratik and Mishra, Bamdev},
  journal={Machine Learning},
  volume={115},
  number={4},
  pages={84},
  year={2026},
  publisher={Springer}
}

@article{lawhern2018eegnet,
  title={{EEGNet}: a compact convolutional neural network for {EEG}-based brain--computer interfaces},
  author={Lawhern, Vernon J and Solon, Amelia J and Waytowich, Nicholas R and Gordon, Stephen M and Hung, Chou P and Lance, Brent J},
  journal={Journal of Neural Engineering},
  volume={15},
  number={5},
  pages={056013},
  year={2018},
  publisher={IOP Publishing}
}

@article{lotte2018review,
  title={A review of classification algorithms for {EEG}-based brain--computer interfaces: a 10 year update},
  author={Lotte, Fabien and Bougrain, Laurent and Cichocki, Andrzej and Clerc, Maureen and Congedo, Marco and Rakotomamonjy, Alain and Yger, Florian},
  journal={Journal of Neural Engineering},
  volume={15},
  number={3},
  pages={031005},
  year={2018},
  publisher={IOP Publishing}
}

@article{carrara2025geometric,
  title={Geometric neural network based on phase space for {BCI}-{EEG} decoding},
  author={Carrara, Igor and Aristimunha, Bruno and Corsi, Marie-Constance and de Camargo, Raphael Y and Chevallier, Sylvain and Papadopoulo, Th{\'e}odore},
  journal={Journal of Neural Engineering},
  volume={22},
  number={1},
  pages={016049},
  year={2025},
  publisher={IOP Publishing}
}

@article{chevallier2024largest,
  title={The largest {EEG}-based {BCI} reproducibility study for open science: the {MOABB} benchmark},
  author={Chevallier, Sylvain and Carrara, Igor and Aristimunha, Bruno and Guetschel, Pierre and Sedlar, Sara and Lopes, Bruna and Velut, S{\'e}bastien and Khazem, Salim and Moreau, Thomas},
  journal={arXiv preprint arXiv:2404.15319},
  year={2024}
}

@incollection{prechelt1998early,
  title={Early stopping---but when?},
  author={Prechelt, Lutz},
  booktitle={Neural Networks: Tricks of the trade},
  pages={55--69},
  year={1998},
  publisher={Springer}
}

@inproceedings{li2021fedbn,
  title={{FedBN}: Federated learning on non-{IID} features via local batch normalization},
  author={Li, Xiaoxiao and Jiang, Meirui and Zhang, Xiaofei and Kamp, Michael and Dou, Qi},
  booktitle={International Conference on Learning Representations (ICLR)},
  year={2021}
}

@article{barachant2012multiclass,
  title={Multiclass brain--computer interface classification by {Riemannian} geometry},
  author={Barachant, Alexandre and Bonnet, St{\'e}phane and Congedo, Marco and Jutten, Christian},
  journal={{IEEE} Transactions on Biomedical Engineering},
  volume={59},
  number={4},
  pages={920--928},
  year={2012},
  publisher={IEEE}
}

@book{absil2008optimization,
  title={Optimization algorithms on matrix manifolds},
  author={Absil, P-A and Mahony, Robert and Sepulchre, Rodolphe},
  year={2008},
  publisher={Princeton University Press}
}

@book{boumal2023introduction,
  title={An introduction to optimization on smooth manifolds},
  author={Boumal, Nicolas},
  year={2023},
  publisher={Cambridge University Press}
}

@inproceedings{huang2017riemannian,
  title={A {Riemannian} network for {SPD} matrix learning},
  author={Huang, Zhiwu and Van Gool, Luc},
  booktitle={Proceedings of the {AAAI} Conference on Artificial Intelligence},
  volume={31},
  year={2017}
}

@inproceedings{kobler2022spd,
  title={{SPD} domain-specific batch normalization to crack interpretable unsupervised domain adaptation in {EEG}},
  author={Kobler, Reinmar J and Hirayama, Jun-ichiro and Zhao, Qibin and Kawanabe, Motoaki},
  booktitle={Advances in Neural Information Processing Systems ({NeurIPS})},
  year={2022}
}

@article{brooks2019riemannian,
  title={{Riemannian} batch normalization for {SPD} neural networks},
  author={Brooks, Daniel and Schwander, Olivier and Barbaresco, Fr{\'e}d{\'e}ric and Schneider, Jean-Yves and Cord, Matthieu},
  journal={Advances in Neural Information Processing Systems},
  volume={32},
  year={2019}
}

@article{bouchard2025beyond,
  title={Beyond {$R$}-barycenters: An effective averaging method on {Stiefel} and {Grassmann} manifolds},
  author={Bouchard, Florent and Laurent, Nils and Said, Salem and Le Bihan, Nicolas},
  journal={{IEEE} Signal Processing Letters},
  volume={32},
  pages={1950--1954},
  year={2025},
  publisher={IEEE}
}

@inproceedings{mcmahan2017communication,
  title={Communication-efficient learning of deep networks from decentralized data},
  author={McMahan, Brendan and Moore, Eider and Ramage, Daniel and Hampson, Seth and y Arcas, Blaise Aguera},
  booktitle={International Conference on Artificial Intelligence and Statistics ({AISTATS})},
  pages={1273--1282},
  year={2017},
  organization={PMLR}
}

@article{li2022federated,
  title={Federated learning on {Riemannian} manifolds},
  author={Li, Jiaxiang and Ma, Shiqian},
  journal={Applied Set-Valued Analysis and Optimization},
  volume={5},
  number={2},
  pages={213--232},
  year={2023}
}

@article{zhang2024nonconvex,
  title={Nonconvex federated learning on compact smooth submanifolds with heterogeneous data},
  author={Zhang, Jiaojiao and Hu, Jiang and So, Anthony Man-Cho and Johansson, Mikael},
  journal={Advances in Neural Information Processing Systems},
  volume={37},
  pages={109817--109844},
  year={2024}
}

@inproceedings{huang2024riemannian,
  title={Riemannian federated learning via averaging gradient streams},
  author={Huang, Zhenwei and Huang, Wen and Jawanpuria, Pratik and Mishra, Bamdev},
  booktitle={International Conference on Learning Representations (ICLR)},
  year={2026}
}

@IEEEtranBSTCTL{IEEEexample:BSTcontrol,
  CTLdash_repeated_names      = "no",
  CTLuse_forced_etal          = "yes",
  CTLmax_names_forced_etal    = "4",
  CTLnames_show_etal          = "1"
}

@inproceedings{li2017second,
  title={Is second-order information helpful for large-scale visual recognition?},
  author={Li, Peihua and Xie, Jiangtao and Wang, Qilong and Zuo, Wangmeng},
  booktitle={Proceedings of the {IEEE} International Conference on Computer Vision ({ICCV})},
  pages={2070--2078},
  year={2017}
}

@article{li2020federated,
  title={Federated optimization in heterogeneous networks},
  author={Li, Tian and Sahu, Anit Kumar and Zaheer, Manzil and Sanjabi, Maziar and Talwalkar, Ameet and Smith, Virginia},
  journal={Proceedings of Machine Learning and Systems},
  volume={2},
  pages={429--450},
  year={2020}
}

@article{yuan2022convergence,
  title={On convergence of {FedProx}: Local dissimilarity invariant bounds, non-smoothness and beyond},
  author={Yuan, Xiaotong and Li, Ping},
  journal={Advances in Neural Information Processing Systems},
  volume={35},
  pages={10752--10765},
  year={2022}
}

@article{jafuno2025classification,
  title={Classification of Buried Objects From Ground Penetrating Radar Images by Using Second-Order Deep Learning Models},
  author={Jafuno, Douba and Mian, Ammar and Ginolhac, Guillaume and Stelzenmuller, Nickolas},
  journal={IEEE Journal of Selected Topics in Applied Earth Observations and Remote Sensing},
  volume={18},
  pages={3185--3197},
  year={2025},
  publisher={IEEE}
}

@article{karcher2014riemannian,
  title={{Riemannian} center of mass and so called {Karcher} mean},
  author={Karcher, Hermann},
  journal={arXiv preprint arXiv:1407.2087},
  year={2014}
}

@inproceedings{karimireddy2020scaffold,
  title={{SCAFFOLD}: Stochastic controlled averaging for federated learning},
  author={Karimireddy, Sai Praneeth and Kale, Satyen and Mohri, Mehryar and Reddi, Sashank and Stich, Sebastian and Suresh, Ananda Theertha},
  booktitle={International Conference on Machine Learning ({ICML})},
  pages={5132--5143},
  year={2020},
  organization={PMLR}
}

@inproceedings{yu2019parallel,
  title={Parallel restarted {SGD} with faster convergence and less communication: Demystifying why model averaging works for deep learning},
  author={Yu, Hao and Yang, Sen and Zhu, Shenghuo},
  booktitle={Proceedings of the {AAAI} Conference on Artificial Intelligence},
  volume={33},
  pages={5693--5700},
  year={2019}
}

@article{ablin2024infeasible,
  title={Infeasible deterministic, stochastic, and variance-reduction algorithms for optimization under orthogonality constraints},
  author={Ablin, Pierre and Vary, Simon and Gao, Bin and Absil, Pierre-Antoine},
  journal={Journal of Machine Learning Research},
  volume={25},
  number={389},
  pages={1--38},
  year={2024}
}

@book{goodfellow2016deep,
  title={Deep learning},
  author={Goodfellow, Ian and Bengio, Yoshua and Courville, Aaron},
  year={2016},
  publisher={MIT Press}
}

@inproceedings{kingma2014adam,
  title={{Adam}: A method for stochastic optimization},
  author={Kingma, Diederik P and Ba, Jimmy},
  booktitle={International Conference on Learning Representations (ICLR)},
  year={2015}
}
